\documentclass{article} 
\usepackage{iclr2024_conference,times}

\usepackage{nameref}
\usepackage{epstopdf}
\usepackage{amsmath,amsfonts}
\usepackage{algorithmic}
\usepackage{algorithm}
\usepackage{array}
\usepackage[caption=false,font=footnotesize]{subfig}
\usepackage{textcomp}
\usepackage{url}
\usepackage{verbatim}
\usepackage{graphicx}
\usepackage{amssymb,amsthm}
\usepackage{booktabs} 
\usepackage{color,xcolor}
\usepackage{amstext,bm,rotating}
\usepackage{multirow}
\usepackage{multicol}
\usepackage{threeparttable}
\usepackage{footnote}
\usepackage[title]{appendix}

\usepackage{amsmath,amsfonts,bm}

\def\eqref#1{equation~\ref{#1}}

\def\1{\bm{1}}

\def\vtheta{{\bm{\theta}}}

\def\ve{{\bm{e}}}

\def\vr{{\bm{r}}}

\def\vt{{\bm{t}}}

\def\vv{{\bm{v}}}
\def\vw{{\bm{w}}}
\def\vx{{\bm{x}}}

\def\vTheta{{\bm{\Theta}}}

\DeclareMathAlphabet{\mathsfit}{\encodingdefault}{\sfdefault}{m}{sl}
\SetMathAlphabet{\mathsfit}{bold}{\encodingdefault}{\sfdefault}{bx}{n}

\newtheorem{theorem}{Theorem}

\newtheorem{corollary}[theorem]{Corollary}

\newcommand{\valpha}{{\bm \alpha}}

\newcommand{\vbeta}{{\bm \beta}}

\newcommand{\vC}{{\bm C}}
\newcommand{\vB}{{\bm B}}

\newcommand{\vD}{{\bm D}}

\newcommand{\vA}{{\bm A}}

\newcommand{\vY}{{\bm Y}}

\newcommand{\vX}{{\bm X}}

\newcommand{\vI}{{\bm I}}
\newcommand{\vK}{{\bm K}}
\newcommand{\vZ}{{\bm Z}}

\usepackage{microtype}
\usepackage{soul}
\usepackage{hyperref}
\usepackage{marginnote}
\usepackage{lipsum}
\usepackage{makeidx} 

\definecolor{red}{rgb}{0.1,0.1,0.8}
\definecolor{blue}{rgb}{0,0,0.8}
\definecolor{green}{rgb}{0,0.4,0}

\newcommand{\change}[2]{}
\newcommand{\lchange}[2]{}

\title{Enhancing Kernel Flexibility via Learning Asymmetric Locally-Adaptive Kernels}

\newcommand{\AuthorNames}{Fan He\textsuperscript{1,2}, Mingzhen He\textsuperscript{2}, Lei Shi\textsuperscript{3}, Xiaolin Huang\textsuperscript{2} and Johan A.K. Suykens\textsuperscript{1}}
\newcommand{\Affiliations}{\textsuperscript{1}Department of Electrical Engineering (ESAT), STADIUS Center for Dynamical Systems, Signal Processing and Data Analytics, KU Leuven, Belgium\\ \textsuperscript{2}Institute of Image Processing and Pattern Recognition, Shanghai Jiao Tong University, China \\ \textsuperscript{3}Shanghai Key Laboratory for Contemporary Applied Mathematics, School of Mathematical Sciences, Fudan University, China }
\newcommand{\Emails}{\texttt{\{fan.he, johan.suykens\}@esat.kuleuven.be}, \texttt{\{mingzhen\_he, xiaolinhuang\}@sjtu.edu.cn}, \texttt{leishi@fudan.edu.cn}}

\iclrfinalcopy 
\begin{document}

\clearpage

\pagenumbering{arabic}
\setcounter{page}{1}
\renewcommand\thefigure{\arabic{figure}} 
\renewcommand\thetable{\arabic{table}} 
\setcounter{figure}{0}
\setcounter{table}{0}

\maketitle
\begin{flushleft}
  {\bf\AuthorNames} \\
  \Affiliations \\
  \Emails \\
\end{flushleft}

\bigskip

\begin{abstract}
The lack of sufficient flexibility is the key bottleneck of kernel-based learning that relies on manually designed, pre-given, and non-trainable kernels.
To enhance kernel flexibility, this paper introduces the concept of Locally-Adaptive-Bandwidths (LAB) as trainable parameters to enhance the Radial Basis Function (RBF) kernel, giving rise to the LAB RBF kernel. 
The parameters in LAB RBF kernels are data-dependent, and its number can increase with the dataset, allowing for better adaptation to diverse data patterns and enhancing the flexibility of the learned function. 
This newfound flexibility also brings challenges, particularly with regards to asymmetry and the need for an efficient learning algorithm. 
To address these challenges, this paper for the first time establishes an asymmetric kernel ridge regression framework and introduces an iterative kernel learning algorithm. 
This novel approach not only reduces the demand for extensive support data but also significantly improves generalization by training bandwidths on the available training data. 
Experimental results on real datasets underscore the remarkable performance of the proposed algorithm, showcasing its superior capability in handling large-scale datasets compared to Nystr\"om approximation-based algorithms. Moreover, it demonstrates a significant improvement in regression accuracy over existing kernel-based learning methods and even surpasses residual neural networks.


\end{abstract}

\section{Introduction}\label{sec: intro}

Kernel methods play a foundational role within the machine learning community, offering a lot of classical non-linear algorithms, including Kernel Ridge Regression (KRR, \cite{vovk2013kernel}), Support Vector Machines (SVM, \cite{cortes1995support}), kernel principal component analysis \citep{scholkopf1997kernel}, and a host of other innovative algorithms.
Nowadays, kernel methods maintain their importance thanks to their interpretability, strong theoretical foundations, and versatility in handling diverse data types \citep{ghorbani2020neural, bach2022information, jerbi2023quantum}. However, as newer techniques like deep learning gain prominence, kernel methods reveal a shortcoming: the learned function's flexibility often falls short of expectations.

Recent studies \citep{Ma2017ThePO, Montanari2020TheIP} demonstrate that the fundamental behavior of a sufficiently flexible model, such as deep models, can interpolate samples while maintaining good generalization ability.
However, interpolation with either single or multi-kernel methods is achieved using kernels close to Dirac function and ridgeless models, leading to large parameter norms and poor generalization.
The reason for imperfect interpolation in kernel-based learning stems from its inherent lack of flexibility.
The flexibility of a model, often referred to as its degree of freedom, is directly indicated by the number of its free parameters.
Recent studies \citep{allen2019convergence, zhou2024learning} have revealed that a model with ample flexibility tends to be over-parameterized.
However, the number of free parameters of classical kernel-based models are constrained by the number of training data points $N$, falling far short of the capabilities observed in over-parameterized deep models.
The primary challenge in augmenting the trainable parameters of kernel methods arises from the reliance on manually designed, fixed kernels in traditional learning algorithms, which are inherently untrainable.

Over the past decade, various data-driven approaches have been explored to introduce trainable parameters to kernels to enhance the flexibility.
However, despite significant progress, there remains ample room for improvement in this area.
For instance, multiple kernel learning (MKL, \cite{gonen2011multiple}) employs linear combinations of kernels instead of a single one, yet the increase in the number of learnable parameters is limited to the number of kernel candidates, falling short of expectations.
Deep kernel learning (\cite{wilson2016deep}) introduces deep architectures as explicit feature mappings for kernels, with the core of its techniques residing in feature learning rather than kernel learning.
Recent works, such as \cite{liu2020learning}, propose the direct introduction of trainable parameters into the kernel matrix but lack corresponding kernel function formulations.
Furthermore, the increased flexibility in these works, while advantageous, has not mitigated the computational issues arising from the large $N \times N$ kernel matrices.

In this paper, we introduce a novel approach to enhance kernel flexibility, significantly differing from current research endeavors. Our proposal involves augmenting the RBF kernel by introducing locally adaptive bandwidths. Given a dataset $\mathcal{X}=\{\vx_1,\cdots,\vx_N\}\subset\mathcal{R}^M$ and let $\odot$ denote the Hadamard product, then the kernel defined on $\mathcal{X}$ is outlined below:
\begin{equation}\label{equ: kernel function}
    \mathcal{K}(\vt,\vx_i) 
    = \exp\left\{- \|\vtheta_i\odot(\vt-\vx_i)\|_2^2\right\},\qquad \forall \vx_i\in\mathcal{X}, \;\forall \vt\in\mathcal{R}^M,
\end{equation}
where $\vtheta_i\in \mathcal{R}_+^M,\forall i$ denotes a positive bandwidth\footnote{Strictly speaking, the bandwidths in a LAB RBF kernel should be a vector function defined on $\mathcal{R}^M$.  However, for better clarity in illustrating the subsequent learning algorithm, we discretely define the bandwidth for each support vector data point in a point by point way.}.
We name (\ref{equ: kernel function}) as Local-Adaptive-Bandwidth RBF (LAB RBF) kernels.
The key difference between LAB RBF kernels and conventional RBF kernels lies in assigning distinct bandwidths $\vtheta_i$ to each sample $\vx_i$ rather than using a uniform bandwidth across all data points, and we proposed to estimate these bandwidths $\vtheta_i$ from training data.

\begin{figure*}[tb]
\begin{center}
\centerline{\includegraphics[width=0.85\textwidth]{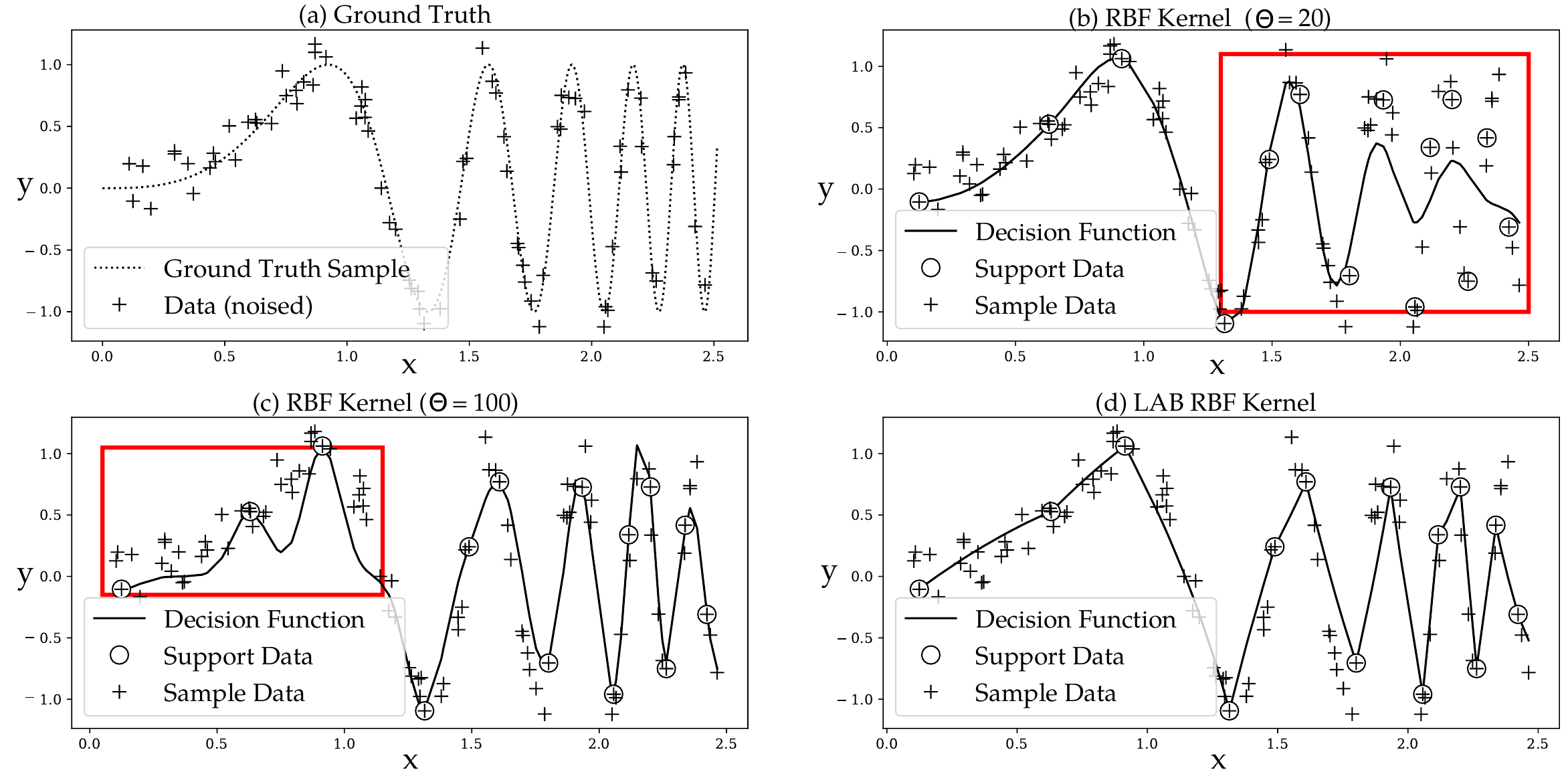}
}
\caption{A toy example demonstrating the regression of a one-dimensional signal $y=\sin(2x^3)$. (a) Ground truth. (b) Function obtained with a universal kernel bandwidth, showing a lack of accuracy in high-frequency regions. (c) Function obtained with a increased bandwidth, resulting in unnecessary sharp changes in low-frequency areas. (d) In our proposed LAB RBF kernels, data-dependent bandwidths are trained, effectively adapting to the underlying function: larger bandwidths correspond to the left part, and smaller bandwidths correspond to the right part.
}
\label{fig: example}
\end{center}
\end{figure*}

Obviously, the introduction of such data-dependent bandwidths can effectively increase the flexibility of kernel-based learning, making bandwidths adaptive to data, not universal as in conventional kernels. Fig.~\ref{fig: example} illustrates the benefits of introducing locally-adaptive bandwidths.
The underlying function $y=\sin(2x^3)$ exhibits varying frequencies across its domain. 
When employing the RBF kernel with a global bandwidth, a dilemma arises: using a small bandwidth (=20) yields an inadequate approximation of the high-frequency portion (highlighted in the red box in Fig.\ref{fig: example} (b)). Conversely, a larger bandwidth (=100) is necessary to approximate the high-frequency section, resulting in a final function that is overly sharp and struggles to accurately represent the smooth portions (highlighted in the red box in Fig.\ref{fig: example} (c)).
The proposed LAB RBF kernel emerges as an optimal solution, offering a more flexible approach to bandwidths. As depicted in Fig.~\ref{fig: example} (d), it strategically employs larger bandwidths on the left and smaller bandwidths on the right.

Notably, LAB RBF kernels exhibit inherent asymmetry since $\vtheta_i$ is not required to be equal to $\vtheta_j$, allowing for cases where $\mathcal{K}(\vx_i,\vx_j)\neq\mathcal{K}(\vx_j,\vx_i)$.
This asymmetry, while providing added flexibility to LAB RBF kernels compared to traditional symmetric kernels, presents two algorithmic challenges: \textit{how to determine the bandwidths for support data?} and \textit{how to incorporate asymmetric LAB RBF kernels into existing kernel-based models?}
This paper addresses these challenges by developing an asymmetric KRR model and introducing an innovative technique for learning asymmetric kernels directly from the training dataset. 
To summarize, the contributions of this paper are as follows:

\textbf{Flexible and trainable kernels.}
The introduced LAB RBF kernel, as presented in (\ref{equ: kernel function}), incorporates individualized bandwidths for each data point, introducing asymmetry and thereby augmenting the number of trainable parameters. 
This augmentation significantly boosts the model's flexibility when employing LAB RBF kernels in kernel-based learning, enabling it to better accommodate a wide array of diverse data patterns.

\textbf{Asymmetric kernel ridge regression framework.} For the application of asymmetric LAB RBF kernels, this paper for the first time establishes an asymmetric KRR framework.
An analytical expression for the stationary points is derived, elegantly represented as a linear combination of function evaluations at training data. 
Remarkably, coefficients of the combination take the same form as those of classical symmetric KRR models, 
despite the asymmetric nature of the kernel matrix.

\textbf{Robust kernel learning algorithm.} We introduce a novel kernel learning algorithm tailored for LAB RBF kernels, enabling the determination of local bandwidths. 
This algorithm empowers the regression function not only to interpolate support data effectively but also to achieve excellent generalization ability by tuning bandwidths on the training data.

Experimental results highlight the advance performance of our algorithm: (1) Regression accuracy is significantly improved compared to current kernel-based methods and surpasses that of residual neural networks in many datasets.
(2) Equipped with a well-learned flexible kernel, the number of support data required in the regression function is significantly reduced, resulting in shorter decision times and better applicability for handling large datasets compared to current kernel-based methods.

\section{Asymmetric Kernel Ridge Regression}
\subsection{Kernel Ridge Regression}
Kernel ridge regression \citep{vovk2013kernel} is one of the most elementary kernelized algorithms.
Define the dataset $\mathcal{X} = \{\vx_1,\cdots,\vx_N\}\subset\mathcal{R}^M, \mathcal{Y}= \{y_1,\cdots,y_N\}\subset\mathcal{R}$, and data matrix $\vX=[\vx_1,\vx_2,\cdots,\vx_N]\in\mathcal{R}^{M\times N}, \vY=[y_1,y_2,\cdots,y_N]^\top\in\mathcal{R}^N$.
The task is to find a linear function in a high dimensional feature space, denoted as $\mathcal{R}^F$, which models the dependencies between the features $\phi(\vx_i), \forall\vx_i\in\mathcal{X}$ of input and
response variables $y_i, \forall y_i\in\mathcal{Y}$.
Here, $\phi: \mathcal{R}^M\to \mathcal{R}^F$ denotes the feature mapping from the data space to the feature space.
Define $\phi(\vX) = [\phi(\vx_1),\phi(\vx_2),\cdots,\phi(\vx_N)]$, then the classical optimization model is as follow:
\begin{equation}\label{equ: skrr}
    \min_{\vw}\;\; \frac{\lambda}{2}\vw^\top \vw + \frac{1}{2}\|\vY - \phi(\vX)^\top \vw\|_2^2,
\end{equation}
where $\lambda>0$ is a trade-off hyper-parameter.
By utilizing the following well-known matrix inversion lemma (see \cite{petersen2008matrix,murphy2012machine} for more information),
\begin{equation}\label{equ: eq}
        (\vA+\vB \vD^{-1}\vC)^{-1}\vB \vD^{-1} = \vA^{-1}\vB(\vC\vA^{-1}\vB +\vD)^{-1},
    \end{equation}
one can obtain the solution of KRR as follow
\begin{equation*}
    \vw^* = (\phi(\vX)\phi(\vX)^\top+\lambda\vI_F)^{-1}\phi(\vX)\vY 
    \overset{(a)}= \phi(\vX)(\lambda\vI_N + \phi(\vX)^\top\phi(\vX))^{-1}\vY,
\end{equation*}
where (\ref{equ: eq}) is applied in (a) with $\vA=\vI_F, \; \vB=\phi(\vX),\; \vC =\phi^\top(\vX),\; \vD=\vI_N$. 

\subsection{Asymmetric Kernel Ridge Regression}
Assume we have two feature mappings from data space to an unknown vector space: $    \phi:\mathcal{R}^M\to \mathcal{R}^F,$ and $    \psi:\mathcal{R}^M\to \mathcal{R}^F.$
Given training dataset $(\vX,\vY)$, the asymmetric kernel ridge regression model is 
\begin{equation}\label{equ: asy krr}
\begin{aligned}
    &\min_{\vw,\vv} \lambda \vw^\top \vv + \frac{1}{2}\|\phi^\top(\vX)\vw-\vY\|_2^2+\frac{1}{2}\|\psi^\top(\vX)\vv-\vY\|_2^2 - \frac{1}{2}\|\psi^\top(\vX) \vv - \phi^\top(\vX) \vw\|_2^2.
\end{aligned}
\end{equation}
Here $\lambda>0$ is again a trade-off hyper-parameter.
Because there are two feature mappings, we have two regressors in the space $\mathcal{R}^F$: $f_1(\vt) = \phi^\top(\vt)\vw$ and $f_2(\vt) = \psi^\top(\vt)\vv$.
The terms $\frac{1}{2}\|\phi^\top(\vX)\vw-\vY\|_2^2+\frac{1}{2}\|\psi^\top(\vX)\vv-\vY\|_2^2$ are used to minimize the regression error.
And the purpose of the terms $\lambda \vw^\top \vv- \frac{1}{2}\|\psi^\top(\vX) \vv - \phi^\top(\vX) \vw\|_2^2 $ is to pursue the substantial distinction between the two regressors.
Then we have the following result on the stationary points.

\begin{theorem}\label{the: solution}
One of the stationary points of (\ref{equ: asy krr}) is
\begin{equation}
    \vw^* = \psi(\vX)(\phi^\top(\vX)\psi(\vX)+\lambda \vI_N)^{-1}\vY, \quad\quad 
    \vv^* = \phi(\vX)(\psi^\top(\vX)\phi(\vX)+\lambda \vI_N)^{-1}\vY.
\end{equation}
\end{theorem}
The proof is presented in Appendix~\ref{apdx: 0}.
Theorem~\ref{the: solution} proves an important conclusion that the stationary points still can be expressed as a linear combination of function evaluations on training dataset, validating the practical feasibility of the proposing framework.
With the conclusion in Theorem~\ref{the: solution}, we can easily apply asymmetric kernel functions.
Define an asymmetric kernel by the inner product of $\phi$ and $\psi$, i.e. $$\mathcal{K}(\vx,\vt) = \langle \phi(\vx), \psi(\vt)\rangle, \;\;\forall \vx,\vt \in \mathcal{R}^M,$$ and denote a kernel matrix $[\vK(\vX,\vX)]_{ij}=\mathcal{K}(\vx_i,\vx_j)=\phi^\top(\vx_i)\psi(\vx_j)$, $\forall \vx_i,\vx_j \in \mathcal{X}$, 
then we obtain two regression functions:
\begin{equation}\label{equ: regressors}
    \begin{aligned}
        &f_1(\vt) = \phi(\vt)^\top\vw^* = \vK(\vt,\vX)(\vK(\vX,\vX)+\lambda\vI_N)^{-1}\vY,\\
        &f_2(\vt) = \psi(\vt)^\top\vv^* = \vK^\top(\vX,\vt)(\vK^\top(\vX,\vX)+\lambda\vI_N)^{-1}\vY.
    \end{aligned}
\end{equation}

Theorem~\ref{the: solution} also indicates the proposed asymmetric KRR framework includes the symmetric one. That is, model~(\ref{equ: asy krr}) and model~(\ref{equ: skrr}) share the same stationary points when the two feature mappings are equivalent, as shown in the following corollary.

\begin{corollary}
If the two feature mappings $\phi$ and $\psi$ are equivalent, i.e. $\phi(\vx) = \psi(\vx), \forall \vx\in\mathcal{R}^M$, then stationary conditions of the asymmetric KRR model (\ref{equ: asy krr}) and the symmetric KRR model (\ref{equ: skrr}) are equivalent. And the stationary point is
$\vw^* =\vv^*=\phi(\vX)(\lambda\vI_N + \phi(\vX)^\top\phi(\vX))^{-1}\vY.$
\end{corollary}

\subsection{Alternative derivation and function explanation}
We can also derive a similar result in Theorem~\ref{the: solution} in a LS-SVM-like approach \citep{suykens1999least}, from which we can better understand the relationship between the two regression functions.
By introducing error variables $e_i = y_i - \phi(\vx_i)^\top\vw$ and $r_i = y_i - \psi(\vx_i)^\top\vv$, the last three terms in (\ref{equ: asy krr}) equals to $\sum_i( e_i^2 +  r_i^2 - (e_i-r_i)^2) $, which further equals to $\sum_i e_ir_i$.
According to this result, we have the following optimization:
\begin{equation}\label{equ: ls akrr}
    \begin{aligned}        \min_{\vw,\vv,\ve,\vr} \;\;&\lambda \vw^\top\vv + \sum_{i=1}^N e_i r_i\\
    \mathrm{s.t.}\;\;& e_i = y_i - \phi(\vx_i)^\top\vw,\quad \forall i = 1,2,\cdots, N,\\
    &r_i = y_i - \psi(\vx_i)^\top\vv,\quad \forall i = 1,2,\cdots, N.
    \end{aligned}
\end{equation}
From the Karush-Kuhn-Tucker (KKT) conditions \citep{boyd2004convex}, we can obtain the following result on the KKT points.
\begin{theorem}\label{the: solution2}
Let $\valpha = [\alpha_1,\cdots,\alpha_N]^\top \in\mathcal{R}^N$ and $\vbeta= [\beta_1,\cdots,\beta_N]^\top \in\mathcal{R}^N$ be Lagrange multipliers of constraints $e_i = y_i - \phi(\vx_i)^\top\vw$ and $r_i = y_i - \psi(\vx_i)^\top\vv, \forall i=1,\cdots,N$, respectively. Then one of the KKT points of (\ref{equ: ls akrr}) is
\begin{align*}
    &\vw^* = \frac{1}{\lambda}\psi(\vX)\vbeta^* , \quad\quad &&\ve^*=\vbeta^*= \lambda(\phi^\top(\vX)\psi(\vX)+\lambda \vI_N)^{-1}\vY, \\
    &\vv^* = \frac{1}{\lambda}\phi(\vX)\valpha^*, \quad\quad &&\vr^*=\valpha^* = \lambda(\psi^\top(\vX)\phi(\vX)+\lambda \vI_N)^{-1}\vY.    
\end{align*}
\end{theorem}
This model shares a close relationship with existing models. For instance, by modifying the regularization term from $\vw^\top\vv$ to $\vw^\top\vw+\vv^\top\vv$ and flipping the sign of $\sum_{i=1}^N e_i r_i$, we arrive at the kernel partial least squares model as outlined in \cite{hoegaerts2004primal}. In the specific case where $\psi=\phi$, its KKT conditions align with those of the LS-SVM setting for ridge regression \citep{Saunders1998RidgeRL,Suykens2002LeastSS}. Furthermore, under the same condition of $\psi=\phi$ and when the regularization parameter is set to zero, it reduces to ordinary least squares regression \citep{hoegaerts2005subset}.

With the aid of error variables $\ve$ and $\vr$, a clearer perspective on the relationship between $f_1$ and $f_2$ emerges. As clarified in Theorem~\ref{the: solution2}, the approximation error on training data is equal to the value of dual variables, a computation facilitated through the kernel trick. Consequently, this reveals that $f_1$ and $f_2$ typically diverge when $\phi$ and $\psi$ are not equal, as they exhibit distinct approximation errors.
A complementary geometric insight arises from the term $\sum_{i=1}^N e_i r_i$ within the objective function. This signifies that, in practice, $f_1$ and $f_2$ tend to approach the target $y$ from opposite directions because the signs in their approximation errors tend to be dissimilar. For practical applications, one may opt for the regression function with the smaller approximation error.

\section{Learning Locally-Adaptive-Bandwidth RBF kernels}
\label{algorithm}

\begin{figure}[tb]
    \begin{center}
    {
    \subfloat[{\footnotesize Flowchart of how we learn kernels.}]{\includegraphics[width=0.40\textwidth]{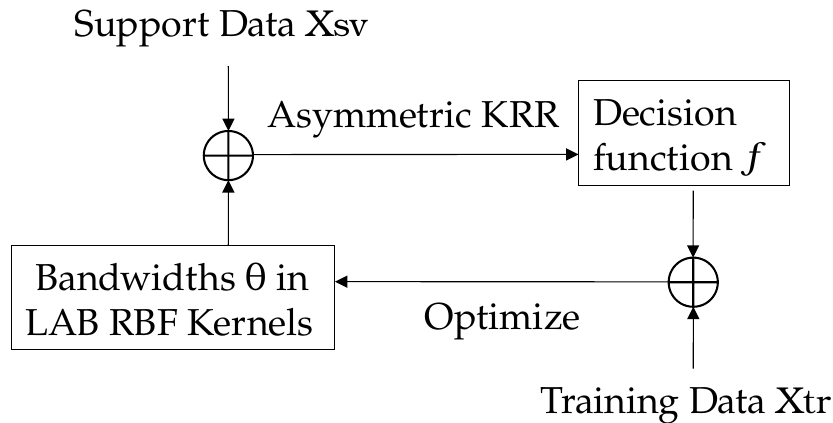}}
    \quad\quad\quad
    \subfloat[{\footnotesize Changes in function spaces when learning kernels.}]{\includegraphics[width=0.50\textwidth]{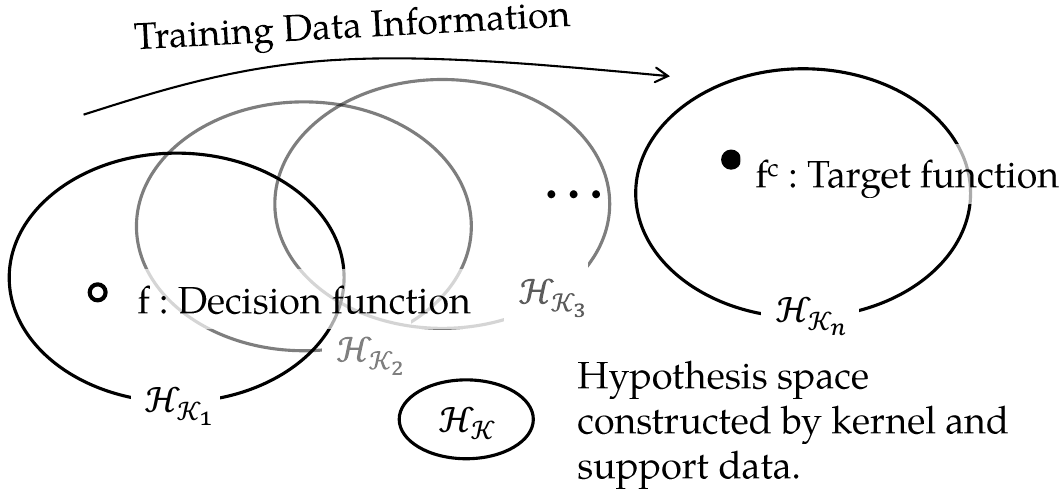}}
    \caption{The idea of the kernel learning in this paper. When interpolating support data to obtain decision function, we actually search in the hypothesis space $\mathcal{H}_\mathcal{K}$. When optimizing $\vTheta$ with training data, we actually adapt the hypothesis space. By repeating these two operation, we finally obtain a good hypothesis space close to the target function.}
    }
    \label{fig: space}
    \end{center}
\end{figure}

Based on the above theoretical result of asymmetric KRR, in this section we provide a learning algorithm of the LAB RBF kernel, to determine the local bandwidths $\vtheta_i$ and  the decision function $f$.
The key of our approach is that, we use a part of available data as \emph{support data}, to which the proposed asymmetric KRR model is applied to obtained the formulation of $f$.
Then we use the rest of data to train bandwidths $\vtheta_i$ corresponding to support data.

The training scheme is given in Fig.~2 (a): Assume a support dataset $\mathcal{Z}_{sv}=\{\mathcal{X}_{sv},\mathcal{Y}_{sv}\}$ and a training dataset $\mathcal{Z}_{tr}=\{\mathcal{X}_{tr},\mathcal{Y}_{tr}\}$ are pre-given.
Let $\vTheta$ denotes the set of bandwidths for support data, i.e., $\vTheta = \{\vtheta_1,\cdots, \vtheta_{N_{sv}}\}$.
We firstly fix $\vTheta$ and apply asymmetric KRR model on support data to obtain the decision function, denoted by $f_{\mathcal{Z}_{sv},\vTheta}$.
Denote the test data as $\vt$, Theorem~\ref{the: solution} and (\ref{equ: regressors}) provides two choices of regression functions associated with $\vK(\vt,\vX_{sv})$ and $\vK^\top(\vX_{sv},\vt)$, respectively. 
It is important to note that within LAB RBF kernels, the matrices $\vK(\vt,\vX_{sv})$ and $\vK^\top(\vX_{sv},\vt)$ are distinct. The bandwidth of the former is dependent on $\vX_{sv}$, while the bandwidth of the latter depends on $\vt$. Since only bandwidths for support data are optimized, we can compute solely $\vK(\vt,\vX_{sv})$, and we are unable to calculate $\vK(\vX_{sv},\vt)$ due to the absence of bandwidth information for testing data.
Consequently, only $f_1$ in (\ref{equ: regressors}) can be utilized to interpolate the support data.
That is,
\begin{equation}\label{equ: interpolation f}
    f_{\mathcal{Z}_{sv},\vTheta}(\vt) 
    = \vK_\vTheta(\vt,\vX_{sv})(\vK_\vTheta(\vX_{sv},\vX_{sv})+\lambda\vI_N)^{-1}\vY_{sv}.
\end{equation}
As a function that interpolates a small dataset, it is evident that the generalization performance of $f_{\mathcal{Z}{sv},\vTheta}$ falls short of expectations. 
However, by introducing trainable bandwidths $\vTheta$, we can substantially improve the generalization capacity of $f_{\mathcal{Z}_{sv},\vTheta}$. This enhancement is achieved by the following optimization model to train  $\vTheta$, enabling the function to approximate the training data:
\begin{equation}\label{equ: opt theta}
\begin{aligned}
    \vTheta^* &= \arg\min_{\vTheta} \|f_{\mathcal{Z}_{sv},\vTheta}(\vX_{tr}) - \vY_{tr}\|_2^2 \\
    &= \arg\min_{\vTheta} \|\vK_\vTheta(\vX_{tr},\vX_{sv})(\vK_\vTheta(\vX_{sv},\vX_{sv})+\lambda\vI_N)^{-1}\vY_{sv}- \vY_{tr}\|_2^2.
\end{aligned}     
\end{equation}
After the above optimization, we finally obtain the regression function $f_{\mathcal{Z}_{sv},\vTheta^*}(\vt) $.

\begin{algorithm}[tb]
    \begin{algorithmic} [1]
    \caption{Learning LAB RBF kernels with SGD and dynamic strategy. }
    \label{alg: AKL}
    \STATE \textbf{Input:} Data $\mathcal{Z}=\{\mathcal{X},\mathcal{Y}\}$, regularization hyper-parameter $\lambda$.  \STATE Initialization: Error tolerance $\epsilon>0$, initial bandwidth $\vTheta^{(0)}>0$, learning rate for gradient descent method $\eta>0$, $k$ for the dynamic strategy, and uniformly sampled support dataset $\mathcal{Z}^{(0)}=\{\mathcal{X}_{sv}^{(0)},\mathcal{Y}_{sv}^{(0)}\}$.  
    \REPEAT
        \STATE Compute the function $ f_{\mathcal{Z}_{sv}^{(t)},\vTheta^{(t)}}$ according to (\ref{equ: kernel function}) and (\ref{equ: interpolation f}).
        \STATE t=0.
    	\REPEAT
    	    \STATE Randomly sample a subset $\{\mathcal{X}_{s},\mathcal{Y}_{s}\}\subset\mathcal{Z}\setminus\mathcal{Z}_{sv}$.
    	    \STATE Compute $\vTheta^{(t+1)}=\vTheta^{(t)}+\eta\frac{\partial}{\partial\vTheta}\mathcal{L}(f_{\mathcal{Z}_{sv}^{(t)},\vTheta^{(t)}}(\vX_{s}),\vY_{s})$ according to (\ref{equ: opt theta}).
            \STATE t=t+1.
        \UNTIL{the maximal number of iteration is exceeded.}
        \STATE Compute error $\xi_i=(f_{\mathcal{Z}_{sv}^{(t)},\vTheta^{(t)}}(\vx_i)-y_i)^2$ for all data $\{\vx_i,y_i\}\in\mathcal{Z}\setminus\mathcal{Z}_{sv}$.
        \IF{$\max_i \xi_i\leq \epsilon$ or the maximal support data number is exceeded,}
        \STATE break.
        \ELSE
        \STATE Add the first $k$ samples with largest error to the support dataset and obtain $\mathcal{Z}_{sv}^{(t+1)}$
        \ENDIF
	\UNTIL{the maximal number of iteration is exceeded.}
    \STATE Compute the $\valpha = (\vK_{\vTheta^{(t)}}(\vX^{(t)}_{sv},\vX_{sv}^{(t)}) + \lambda\vI_N)^{-1}\vY_{sv}^{(t)} $.   
    \STATE \textbf{Return} $\valpha,\mathcal{Z}^{(t)}_{sv}$ and $\vTheta^{(t)}$.
    \end{algorithmic}
\end{algorithm}

In this approach, we utilize support data in $\mathcal{Z}{sv}$ for optimizing the regressor and training data in $\mathcal{Z}{tr}$ for optimizing $\vTheta$. This sets our algorithm apart from existing deep kernel learning and multiple kernel learning methods, where both the decision function and the kernel are optimized concurrently using the same data.
It is worth noting that $\mathcal{Z}_{sv}$ in our strategy serves a similar role to the selected subset of training data in Nystr\"om approximation, (see \cite{williams2000using, rudi2017falkon} for reference). 
However, with the integration of locally-adaptive bandwidths, our subsequent bandwidth training process significantly enhances the generalization ability of decision models, as we will demonstrate in the next experiment section.
Overall, the proposed kernel learning algorithm yields advantages in terms of efficiency and effectiveness:

\textbf{Computational efficiency with small support data.} 
In traditional kernel-based learning, computing the kernel matrix and its inverse is the most computationally demanding operation. However, our approach confines this computation to a small dataset $\mathcal{X}_{sv}$, resulting in a manageable computational complexity of $O(N_{sv}^3)$.
Additionally, in Equation (\ref{equ: opt theta}), the relationship between the objective function and $\vTheta$ is analytically determined, as $f_{\mathcal{Z}{sv},\vTheta}$ is explicitly presented. This enables the utilization of gradient-descent-based methods for training $\vTheta$, leveraging advancements in hardware and software that have significantly improved computational efficiency. Consequently, our algorithm is well-suited for large-scale datasets.

\textbf{Enhanced generalization with large training data.} 
The training procedure for optimizing $\vTheta$ in our approach closely resembles a parameter-tuning step, akin to cross-validation, commonly used to enhance the generalization ability of decision functions in traditional learning frameworks.
Here the generalization ability of $f_{\mathcal{Z}_{sv},\vTheta}$ is also enhanced by effectively approximating a large amount of training data.
This augmentation of generalization stems from the incorporation of training data information into the local bandwidths of LAB RBF kernels. 
These bandwidth adjustments essentially correspond to different hypothesis spaces, as depicted in Fig.~2 (b). By adapting these functional spaces optimally, our approach empowers the regression function to efficiently approximate larger datasets with reduced complexity compared to classical kernel-based models.

\textbf{Dynamic strategy for selecting support data.}
The selection of support data is an important step in constructing a regressor, as it can significantly impact the performance of the final model. There are various strategies for selecting support data, ranging from simple random selection to more sophisticated methods based on label information. For example, sorting data according to their labels and then evenly selecting a required amount of data is a reasonable and practical approach.
Besides, we propose a \textit{dynamic strategy} for selecting support data. Initially, we uniformly select $N_0$ support data points $\mathcal{Z}^{(0)}_{sv}$ and then: 
(i) Optimize (\ref{equ: interpolation f}) and (\ref{equ: opt theta}) accordingly to obtain $f_{\mathcal{Z}^{(0)}_{sv},\vTheta}$.
(ii) Compute approximation error $(f_{\mathcal{Z}^{(0)}_{sv},\vTheta}(\vx_i) - y_i)^2, \forall \{\vx_i,y_i\}\in\mathcal{Z}_{tr}$.
(iii) Add data with first $k$ largest error to form a new support dataset $\mathcal{Z}^{(1)}_{sv}$. 
Repeat the above process until all approximation error is less than a pre-given threshold or the maximal support data number is exceeded.
The overall algorithm is presented in Alg.~\ref{alg: AKL}.
Additionally, the use of gradient methods requires initialization of $\vTheta$, and in our experiments, we suggest using the global bandwidth of a general RBF kernel, tuned on the training data, as the initial parameter.

\section{Numerical Experiments}
\label{sec: exp}
This section is dedicated to assessing the performance of the proposed LAB RBF kernels and Alg.~\ref{alg: AKL} by comparing them to widely used regression methods on real datasets. 
We particularly focus on addressing the following questions: \textit{(1) Does training bandwidths on the training dataset significantly enhance generalization ability? (2) Can the utilization of LAB RBF kernels lead to a reduction in the number of support data and make them applicable to large-scale datasets?}

\subsection{Experiment Setting}

\begin{table}[tb]\scriptsize
\caption{$R^2\;(\uparrow)$  of different regression methods on real datasets.}
\label{tab: exp-1}
\centering
\begin{threeparttable}[b]
\begin{tabular}{c|c|c|c|c|c|c}
\toprule
{\multirow{2}{*}{Dataset}}    & Tecator     & Yacht       & Airfoil   & SML     & Parkinson       & Comp-active\tnote{a}   \\\cline{2-7}
&{N=240,M=122} & {N=308,M=6}  & \multicolumn{1}{c}{N=1503,M=5}  & {N=4137,M=22} & {N=5875,M=20}  & \multicolumn{1}{c}{N=8192,M=21}  \\
\midrule
RBF KRR   & 0.9586$\pm$0.0071  &   0.9889$\pm$0.0025 & 0.8634$\pm$0.0248 &  0.9779$\pm$0.0013 & 0.8919$\pm$0.0091 & 0.9822\\\hline
TL1 KRR   &  0.9670$\pm$0.0113  & 0.9705$\pm$0.0033 & 0.9464$\pm$0.0065  &  0.9947$\pm$0.0005  & 0.9475$\pm$0.0034 & 0.9801 \\\hline
R-SVR-MKL    & 0.9711$\pm$0.0212  &  0.9945$\pm$0.0008  &  0.9201$\pm$0.01099  & 0.9959$\pm$0.0006   &  0.9032$\pm$0.0122  &   0.9834\\\hline
{SVR-MKL}    & 0.9698$\pm$0.0157     &  0.9957$\pm$0.0022   &   0.9535$\pm$0.0042 & 0.9970$\pm$ 0.0006  &  0.9011$\pm$0.0110
  &  0.9829 \\\hline
Falkon    & {0.9769$\pm$0.0086} & 0.9982$\pm$0.0024  & 0.9377
$\pm$0.0067   & 0.9960$\pm$0.0007  & 0.9492$\pm$0.0063 & 0.9808 \\\hline
ResNet    & 0.9841$\pm$0.0067   & 0.9940$\pm$0.0003   &   0.9538$\pm$0.0066  &  0.9976$\pm$0.0004   &  0.9906$\pm$0.0048 &  \textbf{0.9836}  \\\hline
WNN    & \textbf{0.9875$\pm$0.0044} & 0.9924$\pm$0.0025  & 0.9128$\pm$0.0089   & 0.9926$\pm$0.0008  & 0.9139$\pm$0.0055 & 0.9817  \\\hline
{LAB RBF}  &  0.9752$\pm$0.0139 &   \textbf{0.9985$\pm$0.0004}  &   \textbf{0.9608$\pm$0.0079}  & \textbf{0.9990$\pm$0.00005
}   & \textbf{ 0.9950$\pm$0.0009} &  0.9835           \\
\bottomrule
\end{tabular}
    \begin{tablenotes}
        \item [a] The test set of Comp-activ is pre-given. 
    \end{tablenotes}
\end{threeparttable}
\end{table}

\textbf{Datasets.} Real datasets include: Yacht, Airfoil, Parkinson \citep{tsanas2009accurate}, SML, Electrical, Tomshardware from UCI dataset\footnote{\url{https://archive.ics.uci.edu/ml/datasets.php}} \citep{asuncion2007uci}, Tecator from StatLib \footnote{\url{http://lib.stat.cmu.edu/datasets/}} \citep{vlachos2005statlib},Comp-active from Toronto University\footnote{\url{https://www.cs.toronto.edu/~delve/data/comp-activ/desc.html}},  and KC House from Kaggle \footnote{\url{https://www.kaggle.com/datasets/harlfoxem/housesalesprediction}} \citep{harlfoxem2016house}.
The detailed description of datasets are provided in Appendix~\ref{apdx: 2}.
Each feature dimension of data and the label are normalized to $[-1,1]$.

\textbf{Measurement.} 
We use R-squared ($R^2$), also known as the coefficient of determination (refer to \cite{gelman2019r} for more details), on the test set $\mathcal{Z}_{test}$ to evaluate the regression performance.
\begin{displaymath}\small
R^2=1-\frac{\sum_{(\vx_i,y_i)\in \mathcal{Z}_{test}}(y_i-{\hat f(\vx_i)})^2}{\sum_{(\vx_i,y_i)\in \mathcal{Z}_{test}}(y_i-\bar{y})^2},
\end{displaymath}
where $\hat{f}$ is the estimated function and $\bar{y}$ is the mean of labels.
All the following experiments randomly take $80\%$ of the total data as training data and the rest as testing data, and are repeated 50 times.
All the experiments were conducted using Python on a computer equipped with an AMD Ryzen 9 5950X 16-Core 3.40 GHz processor, 64GB RAM, and an NVIDIA GeForce RTX 4060 GPU with 8GB memory.
The code is publicly available at \url{https://github.com/hefansjtu/LABRBF_kernel}.

\textbf{Compared methods.} Six regression methods are compared in this experiment, including:  
\begin{itemize}
    \item RBF KRR \citep{vovk2013kernel}: classical kernel ridge regression with conventional RBF kernels, served as the baseline.
    \item TL1 KRR: classical kernel ridge regression employing an indefinite kernel named Truncated $\ell_1$ kernel \citep{Tl12018Huang}. The expression of TL1 kernel is 
    $     \mathcal{K}(\vx,\vx')=\max\{\rho - \|\vx-\vx'\|_1, 0\},    $
    where $\rho>0$ is a pre-given hyper-parameter. The TL1 kernel is a piecewise linear indefinite kernel and is expected to be more flexible and have better performance than the conventional RBF kernel.
    \item SVR-MKL: Multiple kernel learning applied on support vector regression. The kernel dictionary includes RBF kernels, Laplace kernels, and polynomial kernels. Results for R-SVR-MKL with only RBF kernels are also provided. The implementation of MKL is available in the Python package MKLpy \citep{EasyMKL,lauriola2020mklpy}.
    \item Falkon \citep{rudi2017falkon, meanti2022efficient}: An advanced and well-developed algorithm for KRR that employs hyperparameter tuning techniques to enhance accuracy and utilizes Nyström approximation to reduce the number of support data points, enabling it to handle large-scale datasets. We used the public code of Falkon, available at \url{https://github.com/FalkonML/falkon}.
    \item ResNet: The regression version of ResNet follows the structure in \cite{chen2020deep}, and the code is available in \url{https://github.com/DowellChan/ResNetRegression}.
    \item WNN: The regression version of a wide neural network, which is fully-connected and has only one hidden layer. 
\end{itemize}

All setting and hyper-parameters of these methods are determined for each dataset by 5-fold cross-validation and are given in Appendix~\ref{apdx: 3}.

\begin{table}[tb]\scriptsize
\caption{Number of support vectors of different kernel-based regression methods on real datasets.}
\label{tab: exp-1-sv}
\centering
\begin{tabular}{c|c|c|c|c|c|c}
\toprule
Dataset                                               & Tecator & Yacht & Airfoil & SML  & Parkinson & Comp-active \\ 
\midrule
R-SVR-MKL & 174.2   & 224.7 & 953.1   & 2844 & 4047      & 1397        \\ \hline
SVR-MKL                                               & 160.7   & 144.5 & 1035    & 1424 & 3759      & 1423        \\ \hline
Falkon   & 100   & 200 & 900   & 2000 & 4000     & 1500      \\ \hline
LAB RBF                                               & 20      & 30    & 200     & 340  & 110       & 70          \\ \bottomrule
\end{tabular}
\end{table}

\subsection{Experimental Results}
\textbf{Compared with popular regression methods.}
The results of the regression analysis on small-scale datasets, as measured by the $R^2$, are presented in Table~\ref{tab: exp-1}. It is evident that greater model flexibility leads to improved regression accuracy, thus highlighting the benefits of flexible models. 
Notably, TL1 KRR outperforms RBF KRR in most datasets due to its indefinite nature. R-SVR-MKL, which considers a larger number of RBF kernels, exhibits much better performance than RBF KRR. On the other hand, SVR-MKL, which considers a wider range of kernel types, achieves even higher accuracy compared to R-SVR-MKL. 
Among the neural network models, both ResNet and WNN demonstrate superior performance to the aforementioned methods. Specifically, WNN performs well in datasets with high dimensional features such as Tecator, whereas ResNet excels in larger datasets like Comp-active. Overall, our proposed LAB RBF achieves the highest regression accuracy, significantly increasing the $R^2$ compared to the baseline. Notably, LAB RBF performs better than ResNet in certain datasets, indicating that LAB RBF kernels offer sufficient flexibility and training bandwidths on the training dataset is indeed effective to enhance the model generalization ability.

Table~\ref{tab: exp-1-sv} additionally reports the number of support vectors of sparse kernel methods, which enables a more intuitive understanding of the sizes of decision models. Specifically, it provides the maximal support data number in Alg.~\ref{alg: AKL} for LAB RBF, and the predefined number of centers for Falkon, and the average number of support vectors for SVR methods. 
It should be noted that KRR uses all training data as support data, which results in a much larger complexity of the decision model compared to other kernel methods. 
This observation further underscores the advantage of enhancing kernel flexibility and learning kernels, as demonstrated by the compact size of the decision model achieved with our proposed LAB RBF kernel.

\textbf{Performance on large-scale datasets.}
In Table~\ref{tab: exp-2}, we present a performance comparison between Alg.~\ref{alg: AKL} and three other regression methods on large-scale datasets. 
Traditional KRR algorithms is inefficient on large-scale datasets due to the matrix inverse operator on the large kernel matrix.
Consequently, we compare our algorithm with Falkon, the KRR method employing Nystr\"{o}m approximation to address large-scale datasets. 
The results indicate that LAB RBF kernels exhibit superior performance in terms of both $R^2$ and model complexity.
This is primarily attributed to the significantly higher flexibility of LAB RBF kernels compared to the general kernels used in Falkon. 
Importantly, our results highlight the capability of LAB RBF kernels to effectively handle large-scale datasets while achieving a comparable level of regression accuracy to that of ResNet, a widely acknowledged choice for such datasets. 
Notably, considering the extensive number of parameters in ResNet, it is worth noting that the number of support data for LAB RBF kernels remains relatively low, thanks to the enhanced flexibility offered by locally adaptive bandwidths and the kernel learning algorithm.

\begin{table}[]\scriptsize
\caption{Performance of different algorithms in large-scale datasets. \# S.V. denotes the number of support vectors.}
\label{tab: exp-2}
\centering
\begin{tabular}{c|c|c|c|c|c|c|c|c}
\toprule
\multirow{2}{*}{Dataset}       & \multirow{2}{*}{N}  & \multirow{2}{*}{M}        & WNN  & ResNet  & \multicolumn{2}{c|}{Falkon}  & \multicolumn{2}{c}{LAB-RBF}        \\ \cline{4-9} 
& & &$R^2$ & $R^2$ & {$R^2$} & \# S.V.  & {$R^2$} & \# S.V. \\ 
\midrule 
Electrical    & 10000 & 11                                                  &   0.9617$\pm$0.0027   &    \textbf{0.9705$\pm$0.0025}
 & 0.9532$\pm$0.0025
& 3000& 0.9642$\pm$0.0021     &   300  \\ \hline
KC House    & 21623 & 14                                                 &   0.8501$\pm$0.0194   &    0.8823$\pm$0.0117 & 0.8640$\pm$0.0145 & 5000& \textbf{0.8917$\pm$0.0086}     &   400     \\ \hline
TomsHardware &  28179 & 96 &  0.9248$\pm$0.0303  &   0.9697$\pm$0.0021 
& 0.9001$\pm$0.0143 & 3000  & \textbf{0.9809$\pm$0.0028}      &  500  \\ 
\bottomrule
\end{tabular}
\end{table}

\section{Related Works}
\label{sec: Link}

\textbf{RBF kernels with diverse bandwidths.}
RBF kernels that allow for different bandwidths in local regions have been investigated for a long time in statistics, especially in the fields of kernel regression and kernel density estimation, e.g. \cite{Abramson1982OnBV, Brockmann1993LocallyAB, Zheng2013AdaptivelyWK}.
These pioneering works have demonstrated that locally adaptive bandwidth estimators perform better than global bandwidth estimators in both theory and simulation studies. However, due to limited computing power and problem settings, these works have primarily analyzed one-dimensional algorithms and have not considered the generalization ability. 
In the field of machine learning, works like \cite{steinwart2016learning, hang2021optimal} only propose to employ heterogeneous bandwidths specific to each feature.
However, there has been a limited exploration of data-dependent bandwidths, largely due to a lack of understanding and application of asymmetric kernels. 
In this paper, we address this limitation by introducing an innovative asymmetric KRR framework. Consequently, we achieve locally data-adaptive bandwidths tailored to RBF kernels.

\textbf{Asymmetric kernel learning.}
Existing researches in asymmetric kernel learning primarily focus on the application of asymmetric kernels within established kernel methods \citep{koide2006asymmetric, suykens2016svd, he2023learning} and the interpretation of associated optimization models \citep{lin2022reproducing}. 
Despite notable progress in these contexts, the field still grapples with the scarcity of versatile asymmetric kernel functions. 
Current applications of asymmetric kernel matrices often rely on datasets (e.g. the directed graph in \cite{he2023learning}) or recognized asymmetric similarity measures (e.g. the Kullback-Leibler kernels in \cite{moreno2003kullback}) 
This yields improved performance in specific scenarios but leaving a significant gap in addressing diverse datasets.
With the help of trainable LAB RBF kernels, the framework proposed in this paper thus lays a robust groundwork for harnessing asymmetric kernels in tackling general regression tasks.

\section{Conclusion} \label{sec: clu}
This paper introduces locally-adaptive-bandwidths to RBF kernels, significantly enhancing the flexibility of the resulting LAB RBF kernels.
We tackle the inherent asymmetry of LAB RBF kernels by establishing an asymmetric KRR framework, demonstrating that one of its stationary points maintains a structure akin to classical symmetric KRR. Additionally, an efficient LAB RBF kernel learning algorithm is devised, allowing for bandwidth determination with control over support vectors.
This algorithm can effectively reduce the required number of support data while empowering the regression function to achieve exceptional generalization. Experimental results showcase the algorithm's superiority, surpassing state-of-the-art accuracy and even outperforming well-trained residual networks. This work underscores the benefits of enhancing kernel flexibility and highlights the effectiveness of the proposed asymmetric kernel learning approach.

\subsubsection*{Acknowledgments}
The research leading to these results received funding from the European Research Council under the European Union's Horizon 2020 research and innovation program / ERC Advanced
Grant E-DUALITY (787960). This paper reflects only the authors' views and the Union is not liable for any use that may be made of the contained information.
This work was also supported in part by Research Council KU Leuven: iBOF/23/064; Flemish Government (AI Research Program). Johan Suykens is also affiliated with the KU Leuven Leuven.AI institute.
Additionally, this work received partial support from the National Natural Science Foundation of China under Grants 62376155 and 12171093, as well as from the Shanghai Science and Technology Program under Grants 22511105600, 20JC1412700, and 21JC1400600. Further support was obtained from the Shanghai Municipal Science and Technology Major Project under Grant 2021SHZDZX0102.

\bibliography{thesis}
\bibliographystyle{iclr2024_conference}

\newpage
\begin{appendices}

\section{Proof of Theorem~\ref{the: solution}} \label{apdx: 0}
Here we present the proof of Theorem~\ref{the: solution}.
\begin{proof}
    Based on Equation~(\ref{equ: eq}),    
    we can express $\vw^*$ as:
    \begin{equation}\label{equ: w*}
    \begin{aligned}        
        \vw^* &= \psi(\vX)(\phi^\top(\vX)\psi(\vX)+\lambda \vI_N)^{-1}\vY 
        &\overset{(a)}= (\lambda \vI_F + \psi(\vX)\phi^\top(\vX))^{-1} \psi(\vX)\vY
    \end{aligned}
    \end{equation}
    where equation (a) is derived from (\ref{equ: eq}) with $\vA=\vI_F, \; \vB=\psi(\vX),\; \vC =\phi^\top(\vX),\; \vD=\vI_N$.
    Similarly, for $\vv^*$, we have
    \begin{equation}\label{equ: v*}
        \begin{aligned}
            \vv^* &= \phi(\vX)(\psi^\top(\vX)\phi(\vX)+\lambda \vI_N)^{-1}\vY
            &\overset{(b)}=(\lambda \vI_F + \phi(\vX)\psi^\top(\vX))^{-1} \phi(\vX)\vY,
        \end{aligned}
    \end{equation}
    where equation (b) again applies (\ref{equ: eq}) with $\vA=\vI_F, \; \vB=\phi(\vX),\; \vC =\psi^\top(\vX),\; \vD=\vI_N$.
    Take the derivation of the objective function with respect to $\vw$ and $\vv$ at point $(\vw^*,\vv^*)$, we observe:
    \begin{equation*}
        \begin{aligned}
            \frac{\partial L}{\partial \vw}|_{\vw= \vw^*\atop \vv=\vv^*} &=(\lambda \vI_F + \phi(\vX)\psi^\top(\vX))(\lambda \vI_F + \phi(\vX)\psi^\top(\vX))^{-1} \phi(\vX)\vY -\phi(\vX) \vY =0,\\
            \frac{\partial L}{\partial \vv}|_{\vw= \vw^*\atop \vv=\vv^*}
            &=(\lambda \vI_F + \psi(\vX)\phi^\top(\vX))(\lambda \vI_F + \psi(\vX)\phi^\top(\vX))^{-1} \psi(\vX)\vY -\psi(\vX) \vY=0.
        \end{aligned}
    \end{equation*}
    This verifies that the point $(\vw^*,\vv^*)$ satisfies the stationarity condition.
\end{proof}

\section{Proof of Theorem~\ref{the: solution2}} \label{apdx: 1}
Here we present the proof of Theorem~\ref{the: solution2}.
\begin{proof}

The Lagrangian of (\ref{equ: ls akrr}) is 
\begin{equation}
   \mathcal{L} = \lambda \vw^\top\vv + \sum_{i=1}^N e_i r_i + \sum_i \alpha_i(y_i - e_i -\phi(\vx_i)^\top\vw) + \sum_i \beta_i(y_i - r_i -\psi(\vx_i)^\top\vv),
\end{equation}
where $\valpha\in\mathcal{R}^N$ and $\vbeta\in\mathcal{R}^N$ are Lagrange multipliers.
The KKT conditions lead to
\begin{align*}
    &\frac{\partial \mathcal{L}}{\partial \vv} = \lambda \vw -\psi(\vX)\vbeta = 0 && \Longrightarrow  \vw = \frac{1}{\lambda}\psi(\vX)\vbeta,\\
    &\frac{\partial \mathcal{L}}{\partial \vw} = \lambda \vv -\phi(\vX)\valpha = 0 && \Longrightarrow  \vv = \frac{1}{\lambda}\phi(\vX)\valpha,\\
    &\frac{\partial \mathcal{L}}{\partial r_i} =e_i - \beta_i = 0 && \Longrightarrow  e_i = \beta_i,\\
    & \frac{\partial \mathcal{L}}{\partial e_i}= r_i - \alpha_i = 0 && \Longrightarrow  r_i = \alpha_i,\\
    &\frac{\partial \mathcal{L}}{\partial \alpha_i} =y_i - e_i -\phi(\vx_i)^\top\vw = 0 && \Longrightarrow  e_i = y_i -\phi(\vx_i)^\top\vw,\\
    & \frac{\partial \mathcal{L}}{\partial \beta_i}= y_i - r_i -\psi(\vx_i)^\top\vv = 0 && \Longrightarrow  r_i = y_i - \psi(\vx_i)^\top\vv.\\
\end{align*}
Substitute the first four lines into the last two lines, we can eliminate primal variables $\vw,\vv, \ve,\vr$:
\begin{align*}    
    &\vbeta^* = \vY - \frac{1}{\lambda}\phi(\vX)^\top\psi(\vX)\vbeta^*
    && \Longrightarrow
    \vbeta^* = \lambda(\lambda \vI_N + \phi(\vX)^\top\psi(\vX))^{-1}\vY,\\
    &\valpha^* = \vY - \frac{1}{\lambda}\psi(\vX)^\top\phi(\vX)\valpha^*
    && \Longrightarrow
    \valpha^* = \lambda(\lambda \vI_N + \psi(\vX)^\top\phi(\vX))^{-1}\vY.
\end{align*}
    Thus, we get the result in Theorem~\ref{the: solution2} and the proof is completed.
\end{proof}

\section{Details of datasets and the hyper-parameter setting. } \label{apdx: 2}
\textbf{Tecator}: The objective is to predict the fat content of a meat sample based on its near infrared absorbance spectrum. These measurements are obtained using a Tecator Infratec Food and Feed Analyzer operating in the wavelength range of 850 - 1050 nm through the Near Infrared Transmission (NIT) principle. 
The dataset can be download from \url{http://lib.stat.cmu.edu/datasets/tecator}.

\textbf{Yacht}: The Yacht Hydrodynamics Data Set aims to predict the residuary resistance of sailing yachts based on features such as fundamental hull dimensions and boat velocity. The dataset consists of 308 full-scale experiments conducted at the Delft Ship Hydromechanics Laboratory for this purpose.
The dataset can be download from \url{https://archive.ics.uci.edu/dataset/243/yacht+hydrodynamics}.

\textbf{Airfoil}:
The Airfoil Self-Noise dataset is sourced from NASA and encompasses aerodynamic and acoustic tests on two and three-dimensional airfoil blade sections carried out in an anechoic wind tunnel. The objective is to predict the scaled sound pressure level.
The dataset can be download from \url{https://archive.ics.uci.edu/dataset/291/airfoil+self+noise}.

\textbf{SML}: 
The SML2010 dataset is collected from a monitoring system installed in a domotic house and covers approximately 40 days of monitoring data. The dataset contains missing values, which were imputed using the mean value.
The dataset can be download from \url{https://archive.ics.uci.edu/dataset/274/sml2010}.

\textbf{Parkinson}: 
The Oxford Parkinson's Disease Telemonitoring Dataset comprises various biomedical voice measurements from 42 individuals with early-stage Parkinson's disease enrolled in a six-month trial of a telemonitoring device for remote symptom progression monitoring.
The dataset can be download from \url{https://archive.ics.uci.edu/dataset/189/parkinsons+telemonitoring}.

\textbf{Comp-activ}: The ComputerActivity database records diverse performance metrics, such as bytes read/written from system memory, from a Sun Sparctation 20/712 with 2 CPUs and 128 MB of main memory.
The dataset can be download from \url{https://www.cs.toronto.edu/~delve/data/comp-activ/desc.html}.

\textbf{TomsHardware}: This dataset is a part of Buzz in social media data set, containing examples of buzz events from the social network Tom's Hardware.
The dataset can be download from \url{https://archive.ics.uci.edu/dataset/248/buzz+in+social+media}.

\textbf{KC House}:
The KC House dataset focuses on house prices in King County, encompassing Seattle, and includes homes sold between May 2014 and May 2015.
The dataset can be download from \url{https://www.kaggle.com/datasets/shivachandel/kc-house-data}.

\textbf{Electrical}:  The Electrical dataset pertains to the local stability analysis of a 4-node star system, where the electricity producer is situated at the center. This system implements the Decentral Smart Grid Control concept.
The dataset can be download from \url{https://archive.ics.uci.edu/dataset/471/electrical+grid+stability+simulated+data}.

\section{Details of hyper-parameter setting. } \label{apdx: 3}
\textbf{Implementation details.}
Among the compared methods, Kernel Ridge Regression (KRR) stands as the fundamental technique that combines the Tikhonov regularized model with the kernel trick. The coefficients of kernels for both SVR-MKL and R-SVR-MKL are calculated following the approach in EasyMKL \citep{EasyMKL}. For Falkon, the code is available at \url{https://github.com/FalkonML/falkon}. LAB RBF, ResNet, and WNN are optimized using gradient methods with varying hyper-parameters such as initial points, learning rate, and batch size. The initial weights of both ResNet and WNN are set according to the Kaiming initialization introduced in \cite{He2015DelvingDI}. In the subsequent experiments, the Adam optimizer is initially used, and upon stopping, the SGD optimizer is applied. Early stopping is implemented for the training of ResNet and WNN, where $10\%$ of the training data is sampled to form a validation set, and validation loss is assessed every epoch. The epoch with the best validation loss is selected for testing. Detailed hyper-parameters of all compared methods are provided in Table~\ref{tab: exp-setting} (for small-scale datasets) and Table~\ref{tab: exp-setting-2} (for large-scale datasets).

\begin{figure}
    \centering    \includegraphics[width=0.9\textwidth]{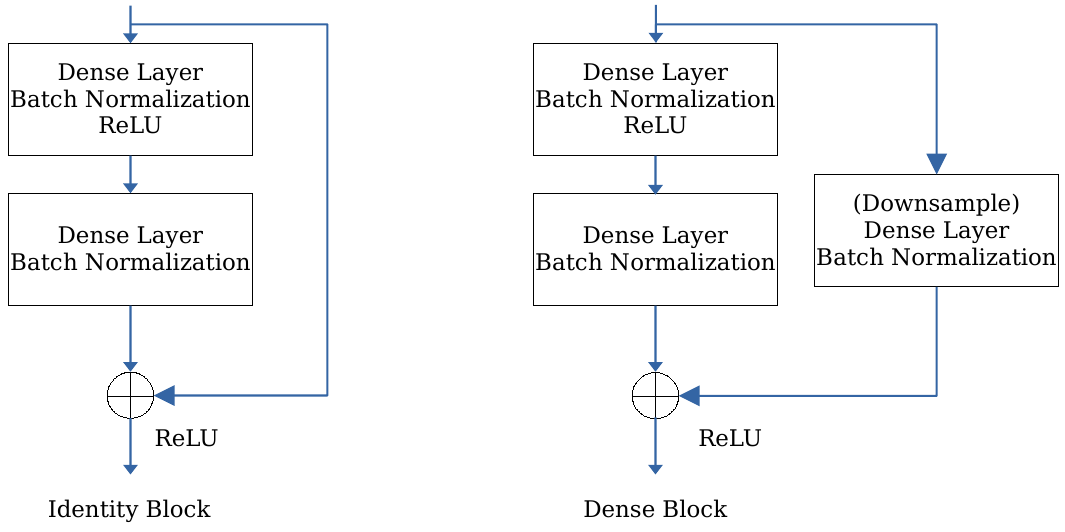}
    \caption{The structures of Identity block and Dense block.}
    \label{fig: resnet}
\end{figure}

The regression version of ResNet follows the structure in \cite{chen2020deep}, which has available code in \url{https://github.com/DowellChan/ResNetRegression}.
Following the structures in \cite{chen2020deep}, the ResNet block has two types: Identity Block (where the dimension of input and output are the same) and Dense Block (where the dimension of input and output are different).
The details of these two block are presented in Fig.~\ref{fig: resnet}.
Considering the different dataset sizes, we use two structures of ResNet in our experiments, denoted by ResNet and ResNetSmall.
For the ResNet, we use two Dense Blocks (M-W-100) and two Identity Block (100-100-100) and a linear predict layer (100-1).
For the ResNetSmall, we use two Dense Blocks (M-W-50)  and a linear predict layer (50-1).
Here W is a pre-given width for the network.

\begin{table}[tbp]\small
    \caption{Hyper-parameters of eight regression methods for real datasets.}
    \label{tab: exp-setting}
    \centering
    \begin{threeparttable}[b]
    \begin{tabular}{c|c|c|c|c|c|c|c}
        \toprule
        &  Hyper-parameters & Tecator & Yacht & Airfoil & SML & Parkinson & Comp\_activ \\ \midrule
        \multirow{3}{*}{LAB RBF} & lr &   0.001 &  0.01  &  0.01  &  0.05   & 0.001 &    0.001 \\ \cline{2-8}
        & Batch size &   16  & 128&   128 &128 &128      &  128        \\ \cline{2-8}
        & $\sigma_0$ &  0.5  &   3  &   10  &  50 &     30   &   0.1     \\ \hline
\multirow{3}{*}{R-SVR-MKL}& C &   1000      &   1000    &     1    &   1000  &    10       &       1      \\ \cline{2-8}
        & $\epsilon$ &    0.001    &   0.001   &    0.01   &   0.001  &   0.01     &    0.01      \\ \cline{2-8} 
        &   Dictionary & \multicolumn{6}{|c} {RBF kernels: $[100,50,10,1,0.1,0.01,0.001]$}
          \\ \hline
\multirow{5}{*}{SVR-MKL}& C &   1000      &   1000    &     100    &   1000  &    1000      &       1000    \\ \cline{2-8}
        & $\epsilon$ &    0.01     &    0.01   &     0.01   &   0.01  &   0.01     &    0.01         \\ \cline{2-8} 
        &  \multirow{3}{*}{Dictionary} & \multicolumn{6}{|c} {RBF kernels: $[100,1,0.1,0.001]$} \\ \cline{3-8} 
        &   & \multicolumn{6}{|c} {Laplace kernels: $[100,1,0.1,0.001]$} \\ \cline{3-8} 
        &   & \multicolumn{6}{|c} {Polynomial kernels: $[1,2,4,10]$} \\ \hline     
\multirow{2}{*}{RBF KRR} & $\sigma$ &    1     &    5   &      80   &  5   &      20     &    10         \\ \cline{2-8}
        & $\lambda$ &     0.01    &  0.001     &   0.001   &   0.01  &   0.001    &    0.001         \\ \hline                                   
\multirow{2}{*}{TL1 KRR }& $\rho$&   98      &    6   &  2.5       &  22   &       14    &     15        \\ \cline{2-8}
        & $\lambda$ &   0.001      &  0.001    &  0.001       &   0.1  &      0.01     &    0.001   \\ \hline
\multirow{3}{*}{Falkon}   & $\lambda$ &      1e-6   &  1e-7     &    1e-6   &   1e-5  &        1e-7   &   1e-6     \\ \cline{2-8}
& Center\tnote{a} &     100    &  200     &    900    &   2000 &   4000    &     1500   \\ \cline{2-8}
& $\sigma$    &     10    &   1    &   2    &  1   &    0.7     &   2.5    \\ \hline
\multirow{3}{*}{ResNet}   & lr &      0.001   &  0.001     &    0.001   &   0.001  &        0.001   &   0.001      \\ \cline{2-8}
        & Batch size &     32    &  32     &    128    &    128&   128    &      128   \\ \cline{2-8}
        & Structure (Width)\tnote{b}    &     2(500)    &   2(500)    &   1(1000)    &  1(1000)   &     1(500)     &   1(2000)    \\ \hline
\multirow{3}{*}{WNN}   & lr &      0.001   &  0.001     &    0.001   &   0.001  &        0.001   &   0.001      \\ \cline{2-8}
        & Batch size &     32    &  32     &    128    &    128&   128    &      128   \\ \cline{2-8}
        & Width    &     800    &   500    &   6000    &  1500   &     3000     &   9000 \\\bottomrule
    \end{tabular}
    \begin{tablenotes}
    \item [a] The center number of Nystr\"om approximation.
        \item [b] Structure 1:$M-W-100-100-100-1$, Structure 2: $M-W-50-1$.
    \end{tablenotes}
\end{threeparttable}
\end{table}

\begin{table}[tbp]\small
    \caption{Hyper-parameters of four regression methods for real datasets.}
    \label{tab: exp-setting-2}
    \centering
    \begin{threeparttable}
    \begin{tabular}{c|c|c|c|c}
        \toprule
        &  Hyper-parameters & TomsHardware  & Electrical & KC House \\ \midrule
        \multirow{3}{*}{LAB RBF} & lr &   0.001 &   0.001  &   0.001  \\ \cline{2-5}
        & Batch size &   256  & 256&   256      \\ \cline{2-5}
        & $\sigma_0$ &  0.1 &  0.1  &   1     \\ \hline
        \multirow{3}{*}{Falkon}   & $\lambda$ &     1e-6  &  1e-6    &    1e-6    \\ \cline{2-5}
        & Center &     3000    &  3000     &    5000    \\ \cline{2-5}
        & $\sigma$    &    2   &   10    &  5  \\ \hline
\multirow{3}{*}{ResNet}   & lr &      0.001   &  0.001     &    0.001     \\ \cline{2-5}
        & Batch size &     128    &  256     &    256    \\ \cline{2-5}
        & Structure (Width)\tnote{a}    &     1(3000)   &   1(2000)    &  1(2000)   \\ \hline
\multirow{3}{*}{WNN}   & lr &      0.01   &  0.01     &    0.01 \\ \cline{2-5}
        & Batch size &     128    &  256     &    128   \\ \cline{2-5}
        & Width    &     3000    &   3000    &   5000   \\\bottomrule
    \end{tabular}
        \begin{tablenotes}
        \item [a] Structure 1:$M-W-100-100-100-1$.
    \end{tablenotes}
\end{threeparttable}
\end{table}

\section{Additional experiment: impact of support data number and initial $\Theta$}
\label{apdx: 4}
In this section, we use synthetic data to evaluate the influence of hyper-parameters in Alg.~\ref{alg: AKL}.
Synthetic data come from typical nonlinear regression test functions provided by \cite{cherkassky1996comparison} and take the following formulations,
\begin{equation*}\small
    \begin{aligned}
        &f_1(\vx)=\frac{1+\sin(2x(1)+3x(2))}{3.5+\sin(x(1)-x(2))},\;D=[-2,-2]^2,\\
        &f_2(\vx)=10\sin(\pi x(1)x(2))+20(x(3)-0.5)^2+5x(4)+10x(5)+0x(6),\;D=[-1,1]^6,\\
        &f_3(\vx)=\exp(2\pi x(1)(\sin(x(4)))+\sin(x(2) x(3))),\;D=[-0.25,0.25]^4,
    \end{aligned}
\end{equation*}
where $D=[a,b]^n=\{\vx|\vx\in \mathcal{R}^n,a\leq x(i)\leq b,\forall 1\leq i \leq n\}$.
The relative sum of square error (RSSE=1-$R^2$) is reported to measure the regression performance.

\begin{figure*}[tb]
\begin{center}
\subfloat[$y=\sin(x^3)$]{\includegraphics[width=0.24\textwidth]{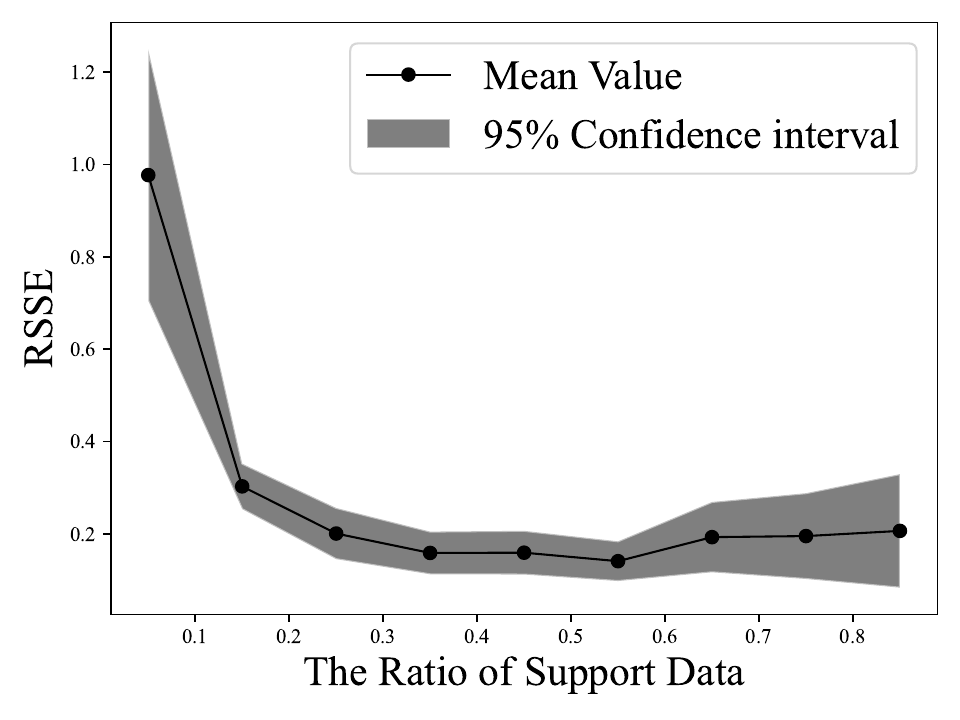}}
\subfloat[$y=f_1(\vx)$]{\includegraphics[width=0.24\textwidth]{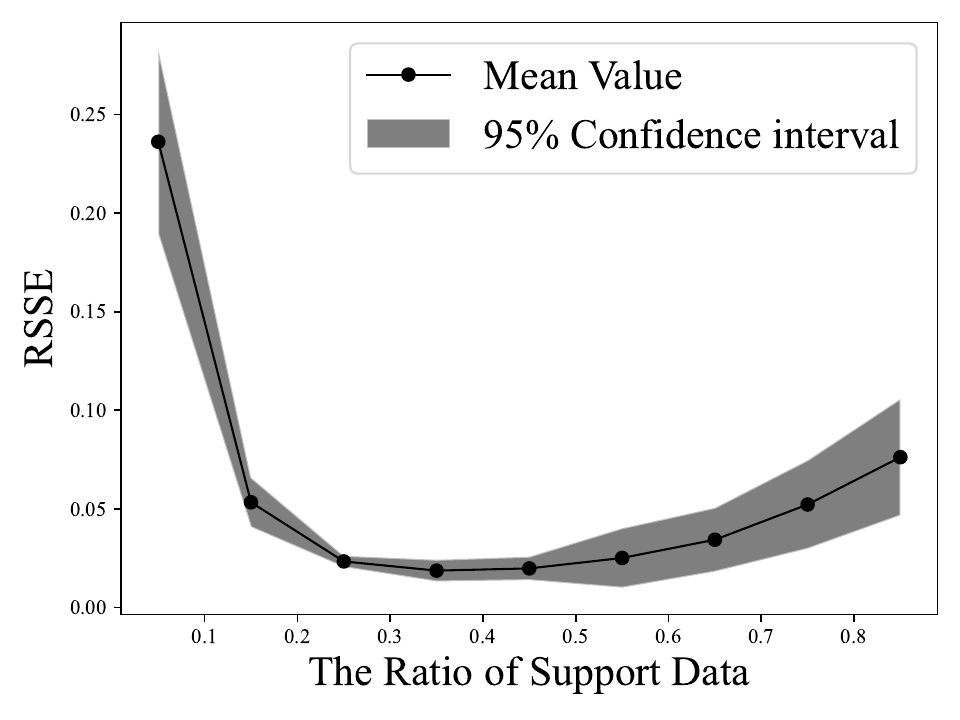}}
\subfloat[$y=f_2(\vx)$]{\includegraphics[width=0.24\textwidth]{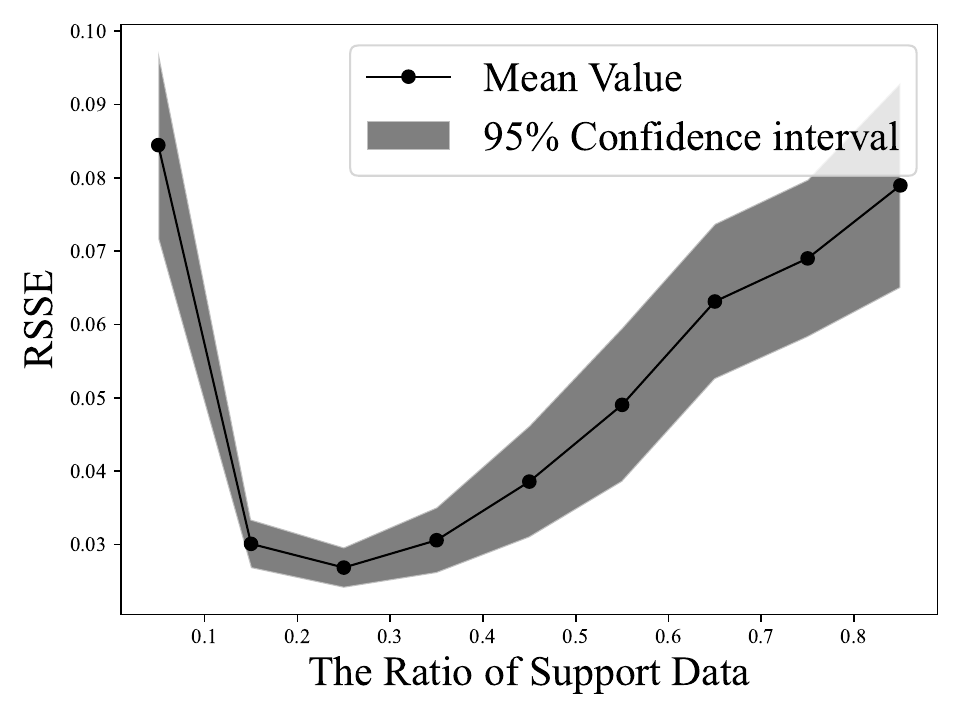}}
\subfloat[$y=f_3(\vx)$]{\includegraphics[width=0.24\textwidth]{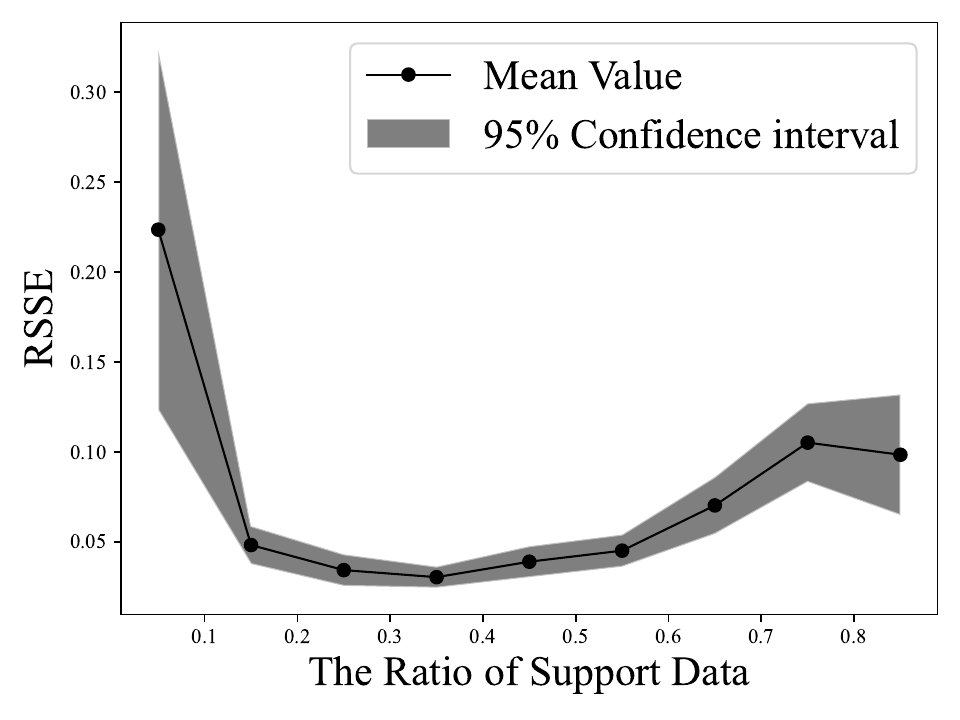}}
\caption{The mean RSSE of Alg.~\ref{alg: AKL} and its standard derivation  with respect to different ratio of support data. $N_{tr}=200, N_{test}=800$. Noise level: (a) $n=0.4$, (b-d) $n=0.1$. }
\label{fig: exp2-ratio}
\end{center}
\end{figure*}

\textbf{The impact of the number of support data.}
The flexibility, or the number of trainable parameters, introduced by LAB RBF kernels is directly controlled by the number of support data in practice\footnote{Note that there is another hyper-parameter in Alg.~\ref{alg: AKL} named error tolerance to control the number of support data. But in practical implementation, setting the maximal support data number has almost equal effect as setting error tolerance, and thus only one of them is considered.}.
To investigate the impact of support data number on the performance of Alg.~\ref{alg: AKL}, we conducted experiments on synthetic datasets with varying ratios of support data, and the results are presented in Figure~\ref{fig: exp2-ratio}. For all functions, we performed random sampling with 200 data points allocated for training and 800 for testing. Initial bandwidths were uniformly set to $\vTheta^{(0)} = 1/M$. We introduced noise at a specified level $n$, with $n$ representing the ratio of noise variance to target variance.

Our findings indicate that Alg.~\ref{alg: AKL} performs optimally when the ratio of support data is in the range of $30\%-70\%$, depending on the feature dimension.
When there are more support data points, the hypothesis space gains greater capacity to capture intricate patterns. However, an excessive number of support data points can lead to overfitting because the remain training data is insufficient, where the model behaves more like simple kernel-based interpolation and becomes sensitive to noise. Conversely, too few support data points may result in underfitting, limiting the model's ability to approximate the data. Thus, selecting an appropriate number of support data points is essential to strike a balance between model complexity and the risk of overfitting.

\begin{figure*}[tb]
\begin{center}
\subfloat[]{\includegraphics[width=0.24\textwidth]{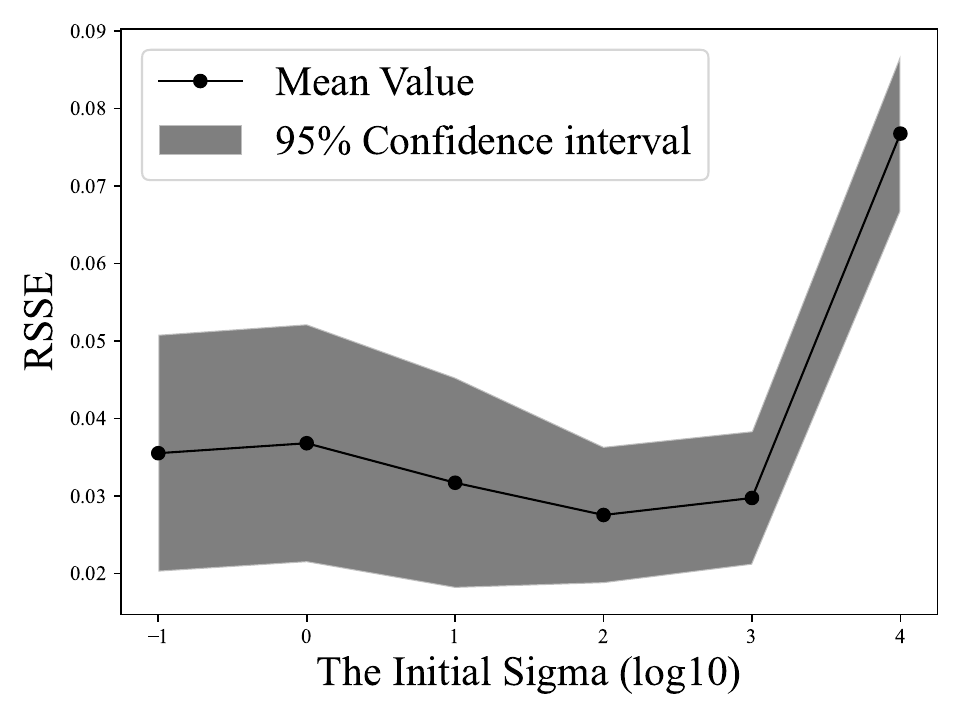}}
\subfloat[]{\includegraphics[width=0.24\textwidth]{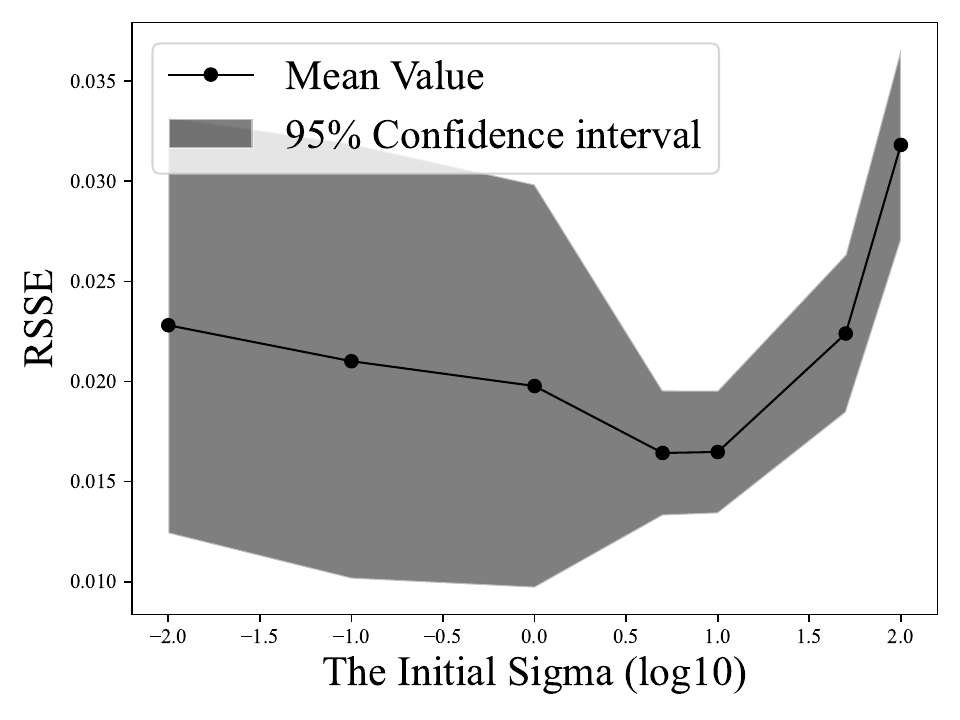}}
\subfloat[]{\includegraphics[width=0.24\textwidth]{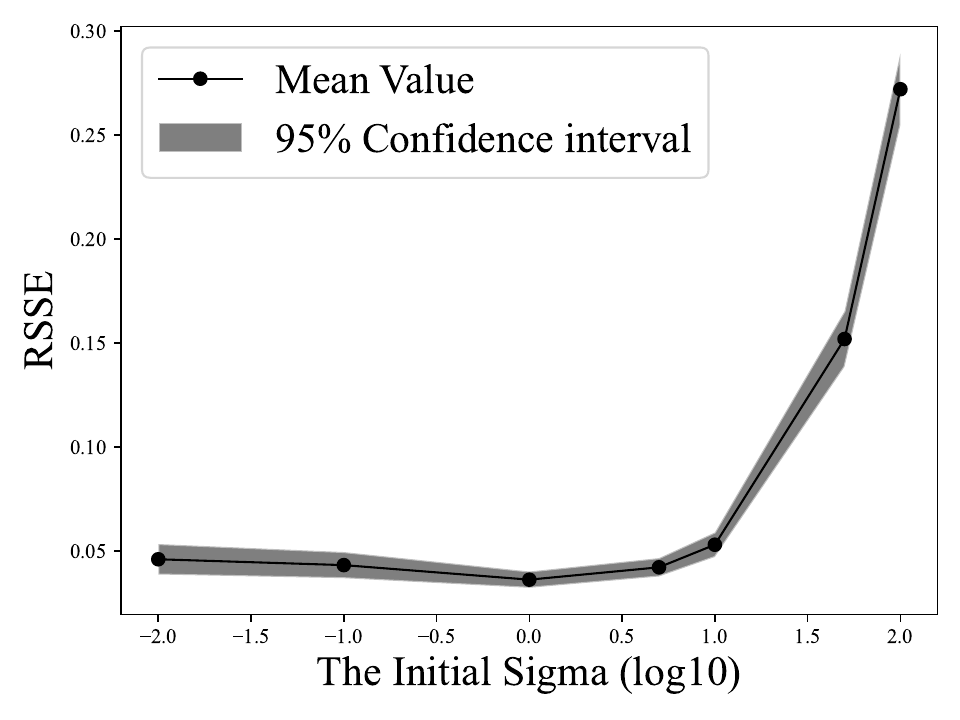}}
\subfloat[]{\includegraphics[width=0.24\textwidth]{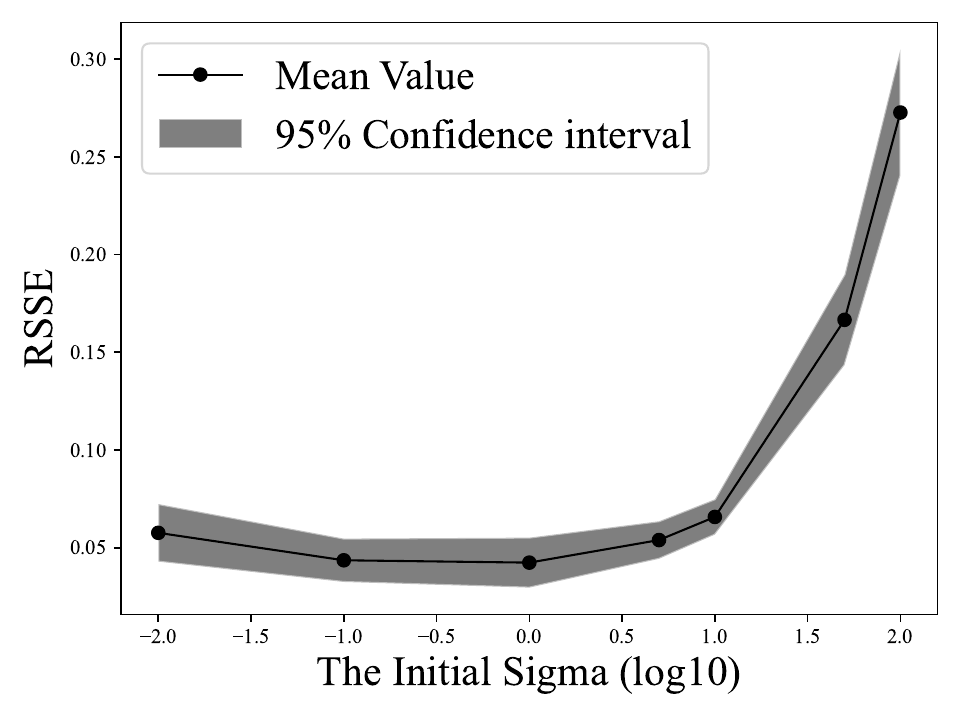}}
\caption{The mean RSSE of Alg.~\ref{alg: AKL} and its standard derivation with respect to different initial $\sigma$. $N_{sv}=100, N_{tr}=100, N_{te}=800$. Datasets: (a) $y=\sin(x^3)$. (b) $y=f_1(\vx)$. (c) $y=f_2(\vx)$.  (d) $y=f_3(\vx)$. Noise level: (a) $n=0.4$, (b-d) $n=0.1$. }
\label{fig: exp3-sigma}
\end{center}
\end{figure*}


\textbf{The impact of the initial parameter $\vTheta^{(0)}$.}
Here, we consider the impact of initial parameter $\vTheta^{(0)}$ because generally, initial points have significant impact on both non-convex optimization and gradient descent method.
Figure~\ref{fig: exp3-sigma} displays the curve of RSSE with respect to different $\vTheta^{(0)}$ indicating that the performance is good and stable in a wide range of $[10^{-2},10]$. 
This result further validates the robustness of our kernel learning algorithm compared to conventional RBF kernels, which are typically sensitive to the pre-given bandwidth.

\end{appendices}

\end{document}